# Sylvester Matrix Based Similarity Estimation Method for Automation of Defect Detection in Textile Fabrics

R.M.L.N. Kumari,[1] G.A.C.T. Bandara,[2] and Maheshi B. Dissanayake [1]

[1] Department of Electrical and Electronics Engineering, University of Peradeniya, KY 20400, Sri Lanka.
[2] Department of Mechanical Engineering, University of Peradeniya, KY 20400, Sri Lanka.

Correspondence should be addressed to Maheshi B. Dissanayake; maheshid@eng.pdn.ac.lk

## Abstract

Fabric defect detection is a crucial quality control step in the textile manufacturing industry. In this article, machine vision system based on the Sylvester Matrix Based Similarity Method (SMBSM) is proposed to automate the defect detection process. The algorithm involves six phases, namely resolution matching, image enhancement using Histogram Specification and Median-Mean Based Sub-Image-Clipped Histogram Equalization, image registration through alignment and hysteresis process, image subtraction, edge detection, and fault detection by means of the rank of the Sylvester matrix. The experimental results demonstrate that the proposed method is robust and yields an accuracy of 93.4%, precision of 95.8%, with 2275 ms computational speed.

## 1. Introduction

Quality is an important aspect in the production line of the textile industry. Thus, fault detection in fabric quality control is an essential requirement of the textile industry. To minimize the manual labor in this endeavor, image analysis and processing techniques are widely used in the industry to automate the defect detection and classification process.

The defects that occur frequently on the fabric pattern, limit the manufacturers who are able to recover only 45-65% of their profits from the off-quality goods [1-2]. Hence, the defect detection process in the textile industry needs to satisfy high expectations of nearly 100% detection accuracy. Therefore, any other methods that are adopted should be able to perform real time defect detection with agility and accuracy. The main challenges encountered include the plethora of types and zones of defects to be detected, as well as the very fine variations that are present between the defects.

In many textile companies, the workers perform the fabric quality control process through human visual examinations. As such quality control is totally observer-dependent it lacks uniformity. Further, the fabric quality control process is highly demanding for a human observer because the type of defects present will vary from fabric to fabric, according to the dynamic nature of the production process.

In this paper, we have generalized and automated fabric quality control, using the Sylvester matrix-based defect detection algorithm, which can easily detect even the very fine defects on the fabrics, by comparing the input image with the reference image. Substantial literature sources are available, related to the algorithmic developments in the textile industry, to detect defects [3-8]. As specified in [9-10], an automated defect detection and classification system will certainly enhance the product quality, and result in heightened productivity. Auto-correlation method is one among the robust algorithms for detecting defects in both patterned and unpatterned fabric [11]. Gabor Wavelet Network (GWN) was chosen as an effective technique to extract texture features from the textile fabrics. Depending upon the features extracted, an optimal Gabor filter was designed for defect detection [12-16]. Reference [17] presents the wavelet sub-window and gray level co-occurrence matrix for defect detection, and the Mahalanobis distance to categorize each wavelet sub-window as either defective or non-defective. Local homogeneity and neural network-based defect detection algorithms are presented in [18]. A design that includes both hardware and software, and which uses the Otsu and Golden image subtraction methods was proposed in [19] to reveal the defects. Its performance on a variety of defects validated the accuracy of the method developed.

In [20] the approach proposed, the fusion analysis for surface detection included a combination of the global and local features for the detection process by extracting and classifying the energy characteristics from the images. Based on the genetic elliptical Gabor filter, a novel method of defect detection was proposed in [21]. After being tuned by the genetic algorithm, the Gabor filter was applied to a variety of samples which show differences in type, shape, size and background.

The Elo rating system was designed to inspect by making fair matches between the partitions from the images [22]. It was estimated to have 97% accuracy with the use of 336 patterned images. The particle analyzer method in [23], reveals higher performance compared to the other traditional methods, as it drives the analysis towards a pre-defined region of interest (ROI), and defines a particle as consisting of a minimum number of pixels. Moreover, a huge number of classes having large intra-class diversity continues to pose a major issue in the Feed Forward Neural Network (FFNN) and Support Vector Machines (SVM)dependent inspection methodologies, as all the classifiers require training regarding the known classes of fabric defects [24-25].

The principal deficits present in the available literature are the overall lower accuracy and the substantial time for decision. The methods described in [26], [27], and [28] are able to achieve accuracy levels of only 90%, 90.6% and 90.8%, respectively. The processing time of the algorithms revealed in [29], [30] stays high at 5.2s and 5.9s, respectively. Furthermore, the methods proposed in [31] and [32] fail to give acceptably accurate performances, while detecting the finer defects in the fabrics.

In this paper, we describe a novel defect detection method which has fast processing and high accuracy, to detect even the very fine defects in the fabrics by comparing the reference and test images. In this method, all the images used are in RGB scale with the identical resolution. First, image enhancement is done on every test image to ensure a better contrast image and thus facilitate defect detection. Later, image registration ensures that all the test images are in proper alignment. After this step, image subtraction is done, to crosscheck the input against the reference image to detect any type of defects. If positive rating is noted for the presence of defects, then edge detection is applied to both the reference and test images, to enable tracing even the finer details. Finally, the Sylvester matrix-based similarity method (SMBSM) is used to identify the defects in the fabrics. The method proposed works with 2275 ms computational speed, and 93.4% average accuracy.

## 2. Proposed Methodology

In this research, an automated fault detection technique is proposed to lessen the degree of human interaction required for fault inspection in fabrics. Three types of fault inspection algorithms exist; namely, the referential, non-referential, and hybrid approaches [33]. The algorithm presented here, is based on the referential approach, in which a reference image is employed to find defects in the test image. The proposed system is depicted in Figure 1.

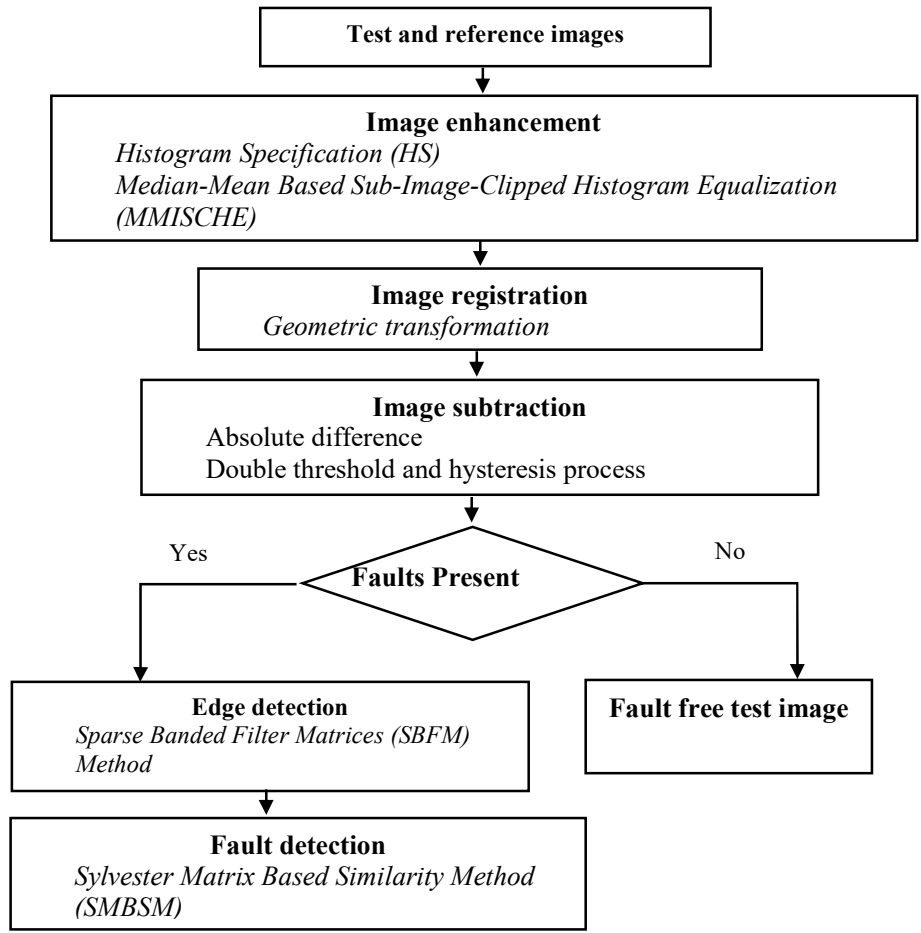

Figure 1: Block diagram of the proposed algorithm.

It is well-known that the performance of any image comparison algorithm is highly dependent upon the capturing condition of the input image. However, our system can analyze the images in the face of different capture conditions, in terms of contrast, distortion and alignments, due to the image pre-processing techniques adopted. In the system proposed, the test image ($I_{Test}$) and the reference image ($I_{Ref}$) are in the RGB format with identical resolution.

### 2.1 Image Enhancement

Image enhancement aims at improving the quality of the test image captured under different lighting conditions. In the algorithm proposed, first the Histogram Specification (HS) improves the contrast level of the test image based on the reference image. Second, the Median-Mean Based Sub-Image-Clipped Histogram Equalization (MMSICHE) algorithm was adopted as the processing technique to achieve the objective of preserving brightness, as well as image information content (entropy) together with control over the enhancement rate. This method circumvents excessive enhancement and provides images having natural enhancement, with the assurance that the test images taken under different lighting conditions will be accurately pre-processed to detect defects.

### 2.1.1 Histogram Specification (HS)

The histogram specifications are used to rectify the contrast levels of the input test image against the reference image. i.e. if the contrast level of the input image is low in comparison to the reference, a correction will be applied to raise the contrast level and vice versa in the event of high contrast inputs [34].

The histogram of the intensity levels of both the reference and test images will be in the range of $[0, L-1]$. Then $n_{Ref}(i)$ and $n_{Test}(i)$ are the number of pixels having intensity $i$ in the reference image and input test image, respectively, where $i = 0, 1, 2, \ldots, L-1$. The inverse transformation, as defined in (1), maps $i$ to $z$ $(0 \leq z \leq L-1)$ and shows the corresponding intensity values of the transformed test image $(G_{T,HS})$.

$$z = \left((L-1)\sum_{z=0}^{i}\frac{n_{Ref}(z)}{p*q}\right)^{-1}\left((L-1)\sum_{j=0}^{i}\frac{n_{Test}(j)}{p*q}\right) \qquad (1)$$

where $p$ and $q$ are the row and column dimension of the images, respectively.

### 2.1.2 Median-Mean Based Sub-Image-Clipped Histogram Equalization (MMISCHE) for contrast enhancement

This method represents the MMISCHE algorithm which consists of three steps: median and mean calculation, histogram clipping and histogram subdivision and equalization. MMISCHE further enhances the image quality of the transformed test image $(G_{T,HS})$.

The median of the image is shown to have an intensity value $X_e$ where the cumulative density function is around 0.5 [35]. Based on the median value, two mean intensity values, the mean of the lower histogram $X_{ml}$ and the mean of the upper histogram $X_{mu}$, are calculated for two individual sub histograms. The corresponding values for the $X_e$, $X_{ml}$ and $X_{mu}$ as shown in Fig. 2, are calculated, as in [35] before the histogram clipping process.

Histogram clipping is done to control the degree of enhancement, to ensure that the resultant image is natural in appearance, matching that of the input image, as close as possible. The clipping threshold $(T_c)$ is calculated as in [35, 28].

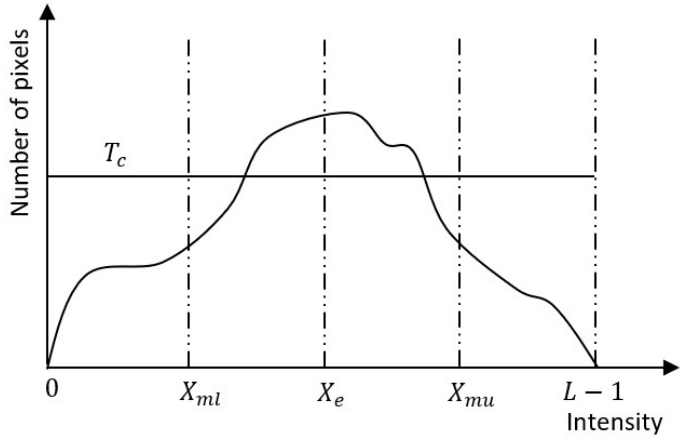

Figure 2: Process of histogram clipping.

The image histogram is divided equally into four bins as shown in Fig. 2. The subdivision process produces four sub images $W_{Ll}, W_{Lu}, W_{Ul}$ and $W_{Uu}$ ranging from the gray level 0 to $X_{ml}$, $X_{ml}+1$ to $X_e$, $X_e + 1$ to $X_{mu}$ and $X_{mu} + 1$ to $L - 1$, respectively. In the next step of MMSICHE, based on the pixel distribution, all the four sub histograms are equalized individually and independently, using either (2), or (3) or (4) or (5), for the independent fine tuning.

$$F_{Ll} = X_{ml} \sum_{i=0}^{X_{ml}} \frac{H_c(i)}{N_{Ll}} \quad \text{for } W_{Ll} \tag{2}$$

$$F_{Lu} = (X_{ml} + 1) + (X_e - (X_{ml} + 1)) \sum_{i=X_{ml}+1}^{X_e} \frac{H_c(i)}{N_{Lu}} \quad \text{for} W_{Lu} \tag{3}$$

$$F_{Ul} = (X_e + 1) + (X_{mu} - (X_e + 1)) \sum_{i=X_e+1}^{X_{mu}} \frac{H_c(i)}{N_{Ul}} \quad \text{for} W_{Ul} \tag{4}$$

$$F_{Uu} = (X_{mu} + 1) + (L - 1 - (X_{mu} + 1)) \sum_{i=X_{mu}+1}^{L-1} \frac{H_c(i)}{N_{Uu}} \text{for } W_{Uu} \tag{5}$$

where $N_{Ll}, N_{Lu}, N_{Ul}$ and $N_{Uu}$ are the total number of pixels in the sub images $W_{Ll}, W_{Lu}, W_{Ul}$ and $W_{Uu}$, respectively.

The final step is to integrate all the sub images, $W_{Ll}, W_{Lu}, W_{Ul}$ and $W_{Uu}$, into one complete image, $(G_{T,MMSICHE})$, for further analysis.

## 2.2 Image Registration

The image registration aims at finding the best transformation, which will align both the reference and input images. More precisely, it is used to identify a correspondence function, or mapping that takes each spatial coordinate from the reference image and returns the coordinate for the test image. The transformation adopted involves two stages, namely the geometric transformation and image re-sampling.

The geometric transformation [37] adopted is expressed as,

$$[x'y'] = [a_{11} a_{12} a_{21} a_{22}][x\ y] + [t_x t_y] \tag{6}$$

where $(x', y')$ is the point coordinate of the test image and $(x, y)$ is the corresponding point coordinate of the reference image. The transformation used in (6) has six degrees of freedom (DOF) where $t_x$ and $t_y$ relate to the translation of the signals, and $a_{11}$, $a_{12}$, $a_{21}$ and $a_{22}$, are used to calculate the scaling and shearing between the two images.

With this transformation, a correspondence map is established between the pixels in the pre-processed test image $(G_{T,MMSICHE})$ and that of the gray scale reference image $(G_R)$, and the registered test image $(G_{T,IR})$ is generated.

## 2.3 Image Subtraction

Image subtraction is done to obtain the differential mapping between the reference image, $(G_R)$ and the preprocessed test image, $(G_{T,IR})$. As image subtraction aims at identifying the presence of defects in the input, it will produce a binary decision which will be "1" if defects are present, and "0" otherwise. If defects are detected, the fabric is then transferred to the edge detection and fault detection stages, where the exact location and details of these defects in the fabric are identified. In the case nothing is detected, the fabric is labeled 'defect-free'.

In the algorithm proposed, the absolute difference $(AD)$ is calculated in a pixel-wise subtraction process, as shown,

$$AD = |G_{T,IR}(x,y) - G_R(x,y)| \qquad (7)$$

The output of $AD$ may include a few erroneous pixels due to the uncorrected noise or misalignment between the two images. The double threshold approach is thus defined as eliminating the non-relevant pixels which belong to the area between $HT_{min}$ and $HT_{max}$, as shown in Fig. 3. The hysteresis process is then performed, where a weak pixel is transformed into a strong one, if and only if at least one strong pixel is present within its neighborhood, as depicted in Fig. 3.

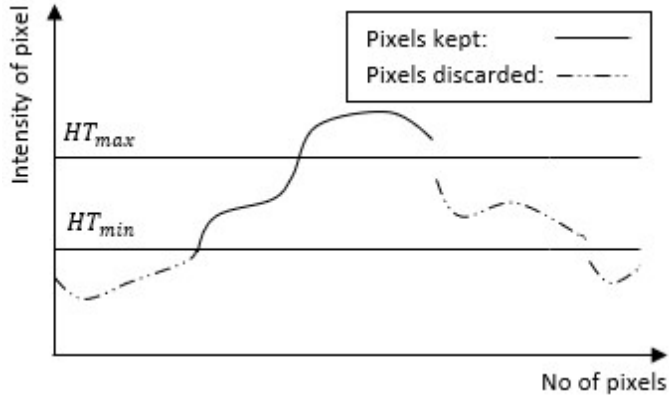

Figure 3: Hysteresis process.

## 2.4 Sparse Banded Filter Matrices (SBFM) for Edge Detection

In the algorithm proposed, Sparse Banded Filter Matrices (SBFM) [38] enables the detection of the edge information in both the test and reference images. The SBFM comprises two major stages, namely implementation of the zero-phase high-pass Butterworth filter using the

SBFM matrix, and edge extraction. This edge detection method facilitates the finer details to be detected, of both the gray scale reference image $(G_R)$ and registered test image $(G_{T,IR})$ significantly, thus ensuring the detection of even considerably insignificant defects on the test image during the fault detection stage.

Implementation of the zero-phase non-causal recursive high-pass filters based on banded matrices was introduced in [38] to identify the edge information from the images.

The matrix form of the first-order Butterworth high-pass filter can be expressed as,

$$y = A^{-1}Bx \qquad (8)$$

where $A$ and $B$ are the banded matrices of size $(N-1) \times (N-1)$ and $(N-1) \times N$, respectively with $N$ representing the length of the input signal. $A$ and $B$ are defined as,

$$A = \begin{bmatrix} a_0 & 0 & . & 0 & 0 & 0 \\ a_1 & a_0 & . & 0 & 0 & 0 \\ 0 & a_1 & . & 0 & 0 & 0 \\ . & . & . & . & . & . \\ 0 & 0 & . & a_1 & a_0 & 0 \\ 0 & 0 & . & 0 & a_1 & a_0 \end{bmatrix}$$

$$B = \begin{bmatrix} -1 & 1 & . & 0 & 0 & 0 & 0 \\ 0 & -1 & . & 0 & 0 & 0 & 0 \\ 0 & 0 & . & 0 & 0 & 0 & 0 \\ . & . & . & . & . & . & . \\ 0 & 0 & . & 0 & -1 & 1 & 0 \\ 0 & 0 & . & 0 & 0 & -1 & 1 \end{bmatrix}$$

Furthermore, the transfer function of the zero-phase non-causal higher-order high-pass Butterworth filter can be expressed as,

$$h(z) = \frac{B(z)}{A(z)} = 1 - \frac{\alpha(-z+2-z^{-1})^d}{(-z+2-z^{-1})^d + \alpha(z+2+z^{-1})^d}$$
$$\alpha = \left(\frac{1-\cos\omega_c}{1+\cos\omega_c}\right)^d \qquad (9)$$

where $d$ and $\omega_c$ are the filter order and a cut-off frequency, respectively.

According to Eq (9), the frequency response is maximally flat at $\omega = 0$, and the frequency response is of unity gain at $\omega = \pi$. Therefore, this is a zero-phase digital filter. The zero-phase high-pass Butterworth filter shown in (9) can be implemented using the (8). Then, $A$ and $B$ can be defined as the banded sparse matrices of size $(N + 2d - 1) \times (N + 2d - 1)$ and $(N + 2d - 1) \times (N + 2d)$, respectively.

The sparse banded high-pass filter proposed is then applied row-wise and column-wise to extract the vertical and horizontal edges, respectively, as in [39], to detect all the edges of the test image processed($G_{T,MMSICHE}$) and the gray scale reference image ($G_R$), while producing their corresponding edge extractions as $G_{T,SBFM}$ and $G_{R,SBFM}$, respectively.

**2.5 Sylvester Matrix Based Similarity Method (SMBSM) for Fault Detection**

The Sylvester matrix ($S$) is associated with two univariate polynomials with the coefficients in a commutative ring [40]. This matrix helps to determine the common roots of the characteristic polynomial of the two images being compared. Hence, the similarity measure between the two images represents the rank or nullity of the matrix $S$.

The $C(i,j)$ and $D(i,j)$ are 2-D sub-images of $G_{R,SBFM}$ and $G_{T,SBFM}$ such that $C(i,j)$, $D(i,j) \epsilon R^{n \times n} (n \leq p, q)$ and square matrices. Their characteristic polynomials can be obtained by evaluating $det\,(\lambda I - C)$ and $det\,(\lambda I - D)$. These characteristic polynomials can be stated as in,

$$P(C) = \sum_{i=0}^{n} c_i \lambda^{n-i} \quad (10)$$

and

$$P(D) = \sum_{i=0}^{n} d_i \lambda^{n-i} \quad (11)$$

The Sylvester matrix $S(E, F) \epsilon R^{(n+n) \times (n+n)}$ of $P(C)$ and $P(D)$, can be defined as,

$$S(E, F) = [E(P(C)) F(P(D))] \quad (12)$$

where $E(P(C)) \epsilon R^{(n+4) \times n}$ and $F(P(D)) \epsilon R^{(n+4) \times n}$ are the Toeplitz matrices whose entries are the coefficients of the $P(C)$ and $P(D)$, respectively [41] and can be defined as,

$$E(P(C)) = \begin{bmatrix} c_0 & 0 & . & 0 & 0 \\ c_1 & c_0 & . & 0 & 0 \\ . & c_1 & . & 0 & 0 \\ c_{n-1} & . & . & c_0 & 0 \\ c_n & c_{n-1} & . & c_1 & c_0 \\ 0 & c_n & . & . & c_1 \\ 0 & 0 & . & c_{n-1} & . \\ 0 & 0 & . & c_n & c_{n-1} \\ 0 & 0 & . & 0 & c_n \end{bmatrix} \quad (13)$$

$$F(P(D)) = \begin{bmatrix} d_0 & 0 & . & 0 & 0 \\ d_1 & d_0 & . & 0 & 0 \\ . & d_1 & . & 0 & 0 \\ d_{n-1} & . & . & d_0 & 0 \\ d_n & d_{n-1} & . & d_1 & d_0 \\ 0 & d_n & . & . & d_1 \\ 0 & 0 & . & d_{n-1} & . \\ 0 & 0 & . & d_n & d_{n-1} \\ 0 & 0 & . & 0 & d_n \end{bmatrix} \qquad (14)$$

The nullity and rank of the matrix $S(E,F)$ show the degree of closeness of the characteristics of $P(C)$ and $P(D)$. For similar images, the nullity value $N(S(E,F))$ is equal to the number of columns in a matrix $S(E,F)$ and is zero for the totally dissimilar images. However, the value of rank, $r(S(E,F))$, is zero for similar images and equal to the number of the columns in a matrix $S(E,F)$ for those images that are totally dissimilar. In the event of small defects, the rank $r(S(E,F))$ too will be small, and the number is seen to rise as the defect intensity increases. Thus, rank $r(S(E,F))$ can be used in a defect intensity function to visualize the defective region, as it is in direct proportion to the defect intensity. Hence, to reach the final labeling in our method we adopt the rank $r(S(E,F))$.

## Results and Discussion

In this section, the findings of the simulation are presented. The algorithm proposed exhibits a significant improvement over the existing methods for defect detection in fabrics, as it is successful in identifying directional defects, under varying conditions of illumination.

The model was assessed in terms of robustness and stability using two datasets KTH-TIPS-I and KTH-TIPS-II [42]. The fabric dataset includes around 500 samples, captured under different conditions of illumination and contrast settings with skews, creating a challenge for defect detection. Table1 lists three samples, including the reference (R) and test (T) images of the dataset. Further, right at the beginning, all the input images are resized to 1024*1024.

First, the image enhancement method described in Section II.A is applied to enhance the histogram of the test image, as shown in Table 2. The intermediate outputs, post application of the image enhancement technique, are depicted in Table 3. Next, the image registration process presented in Section II is applied to align the test image coordinates with the corresponding reference image alignments, as revealed in Table 3. This preprocessing is done to circumvent any inaccuracies at the image subtraction stage, in which the double threshold values $HT_{min}$= 0.035 and $HT_{max}$ = 0.150 are utilized, after manual tuning of the algorithm.

In the textile industry, all the defects need to be detected with 100% accuracy. Keeping this objective in focus, the first priority is to minimize the rate of the false-positives which should be as negligible as possible. If false-negatives occur, then the fabric needs to be re-inspected or discarded, even if the product is defect-free. In this experiment, the rates of the false-positives and false-negatives for the two datasets considered, KTH-TIPS-I and KTH-TIPS-II, are 4.2 % and 0% respectively. This occurred because of the carefully fine-tuned double threshold values used during the subtraction stage. After being subjected to image subtraction, all of the test images (I1T, I2T, and I3T) are identified as defective fabrics. Hence, these images are moved on to the next stage namely, edge detection.

Table 1: The selected reference and test images of the dataset.

| Image ID | Reference image (R) | Test image (T) |
|---|---|---|
| I1 | 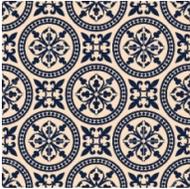 | 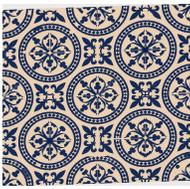 |
| I2 | 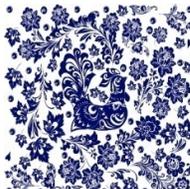 | 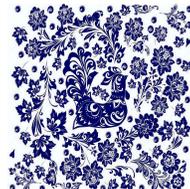 |
| I3 | 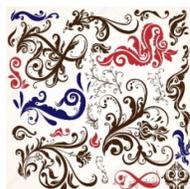 | 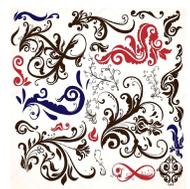 |

In the final stage of the algorithm, edge detection, as shown in Table 4, is performed using SBFM for the reference and test images, which are identified as positively defective, during the image subtraction process. The advantage of having this stage is to enhance all the minor details of the images, which will assist in detecting even the very fine defects present in the fabric.

The input parameters, selected by manual tuning, for the sparse banded high-pass filter design, including degree, cut-off frequency and the length of the input sequence are set as 3, 0.9, and 1024, respectively. The input image size of 1024*1024 is subjected to zero padding, in order to match the sequence length (each row/ column of the input image) given as the input to the filter design. From the analysis it is clear that the edge extraction using SBFM provides more detailed results in the test and reference images, even the finer details, and the discontinuity in the edges extracted is less because the input parameters for this filter design are well tuned and matched. This is evident in the images, as revealed in Table 4.

A comparison of the similarity between the two images is performed using SMBSM. The SMBSM is evaluated using window sizes of 4*4, 8*8, and 12*12 pixels, with the minimum simulation time for the fault detection process being given by the 8*8 window size. The reference and test images are first divided into small sub-windows of 8*8 and then compared against the coinciding location of the reference image. For each sub-window, the Sylvester matrix ($S$) is computed, and its rank is used to determine the defects on the selected sub-window of the test image compared to the reference image. The process is repeated for all the pixels of the entire image to detect faults, as shown in Table 4.

Table 2: histogram of test images after image enhancement.

| Image ID | Histogram of test image before image enhancement | Histogram of test image after image enhancement |
|---|---|---|
| I1 | 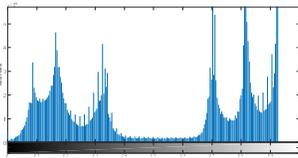 | 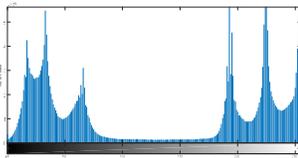 |
| I2 | 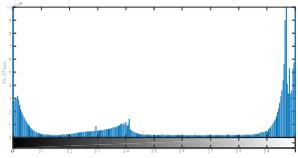 | 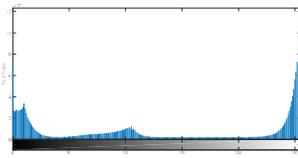 |
| I3 | 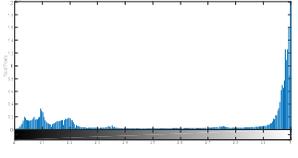 | 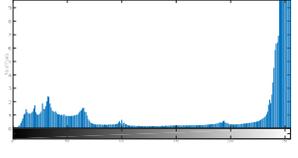 |

Table 3: Test images after image enhancement, coordinate matching and image registration.

| Image ID | Test image after image enhancement | Coordinates matching between reference and test image | After image registration |
|---|---|---|---|
| I1T | 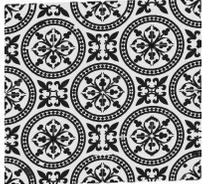 | 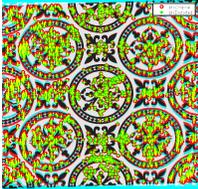 | 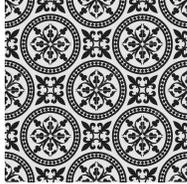 |
| I2T | 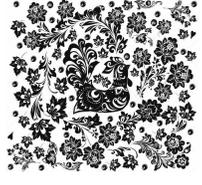 | 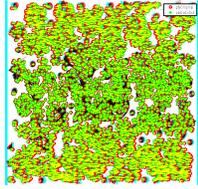 | 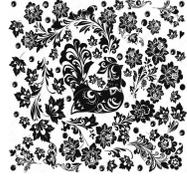 |
| I3T | 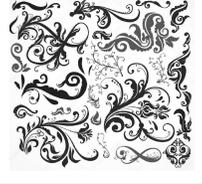 | 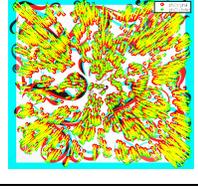 | 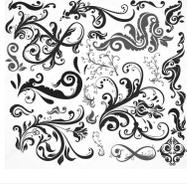 |

Table 4: Edge detection of reference and test image, and fault detection of test image.

| Image ID | Edge detection of reference image | Edge detection of test image | Fault detection on test image |
|---|---|---|---|
| I1 | 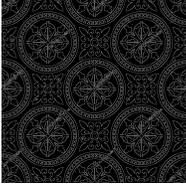 | 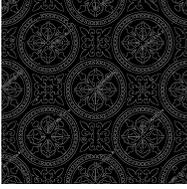 | 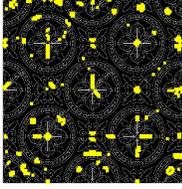 |
| I2 | 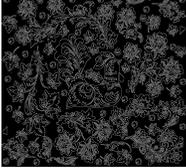 | 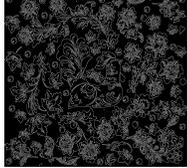 | 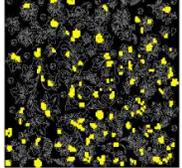 |
| I3 | 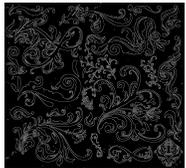 | 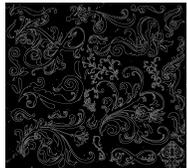 | 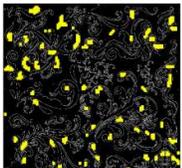 |

To calculate the accuracy of the algorithm we proposed, the Binary Similarity Measure ($\gamma$) is used as given in (15), which detects the dissimilarities between two binary images based on a modified Hamming Distance measure [43]. The values of $\gamma$ range from 0, distinct-dissimilarity, to 1, perfect similarity. The actual and detected faults on the test image are represented in the binary scale $B_{AF}$ and $B_{DF}$, respectively, and the $\gamma$ values of the test images are listed in Table 5.

$$\gamma = \left| 1 - \frac{2}{pq} \sum_{i=1}^{p} \sum_{j=1}^{q} (B_{AF}(i,j) \oplus B_{DF}(i,j)) \right| \quad (15)$$

where the $\oplus$ symbol represents the logical exclusive-or operator, and $p$ and $q$ are the row and column dimensions of the binary images of the actual and detected faults ($B_{AF}$ and $B_{DF}$).

The computational speed of the algorithm proposed is about $2275\ ms$, making it superior to the existing ones. Furthermore, our algorithm performed with 93.4% of accuracy, 95.8% of precision, and 100% of recall, on average. The proposed method works significantly well, even when the test images are taken under different conditions of illumination and have skews. The experiments presented demonstrate the superiority of the proposed method.

Table 5: Actual and detected faults on the test image presented in binary scale.

| Image ID | Actual faults on test image presented in binary scale ($B_{AF}$) | Detected faults on test image presented in binary scale ($B_{DF}$) | Binary Similarity Measure ($\gamma$) |
|---|---|---|---|
| I1 | 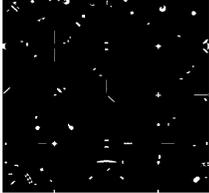 | 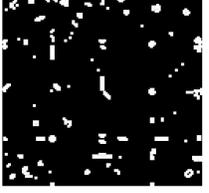 | 0.935 |
| I2 | 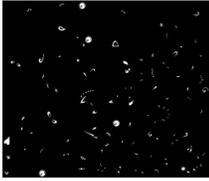 | 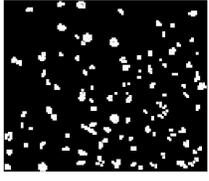 | 0.932 |
| I3 | 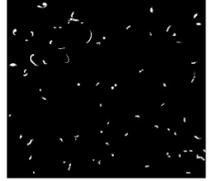 | 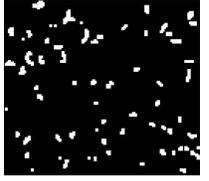 | 0.937 |

## Conclusions

In this paper, a method to identify defects in fabric has been proposed, based on the Sylvester Matrix Based Similarity Method (SMBSM). This method is capable of handling misalignment and varying illuminations of the test images, captured under different conditions, as image enhancement improves the quality of the test image and image registration ensures proper alignment between the reference and test images. Edge detection is guaranteed to identify even very fine defects on fabrics during fault detection. Visual and quantitative results on the two datasets presented have demonstrated that the proposed method is superior and robust. In the future, more experiments will be conducted to further improve the accuracy of this method, and to assure that it is fast enough for defect detection in real time.

## Data Availability

The dataset used in this research is freely available through "M. Fritz, B.C. E. Hayman, and J.O. Eklundh, THE KTH-TIPS Database.
[Online]Available: https://www.nada.kth.se/cvap/databases/kth-tips-I&II. "

## Conflicts of Interest

The authors declare that there is no conflict of interest.